\documentclass{article}

\pdfoutput=1

\usepackage[preprint]{neurips_2019}

\usepackage[utf8]{inputenc} 
\usepackage[T1]{fontenc}    
\usepackage{hyperref}       
\usepackage{url}            
\usepackage{booktabs}       
\usepackage{amsfonts}       
\usepackage{nicefrac}       
\usepackage{microtype}      
\usepackage[pdftex]{graphicx}

\title{NTP : A Neural Network Topology Profiler}

\author{%
  Raghavendra Bhat\\
  Intel Technologies India Pvt Ltd\\
  Bangalore, KA\\
  India, 560103 \\
  \texttt{raghavendra.bhat@intel.com} \\
   \And
 Pravin Chandran \\
  Intel Technologies India Pvt Ltd\\
  Bangalore, KA\\
  India, 560103 \\
  \texttt{pravin.chandran@intel.com} \\
   \And
 Juby Jose \\
  Intel Technologies India Pvt Ltd\\
  Bangalore, KA\\
  India, 560103 \\
  \texttt{juby.jose@intel.com} \\
  \And
 Viswanath Dibbur \\
  Ex-Intel \\
  Bangalore, KA\\
  India, 560103 \\
  \texttt{vdibbur@gmail.com} \\
  \And
 Prakash Sirra Ajith \\
  Ex-Saksen \\
  Bangalore, KA\\
  India, 560103 \\
  \texttt{ajithprakash77@gmail.com} \\
}

\begin{document}

\maketitle

\begin{abstract}
Performance of end-to-end neural networks on a given hardware platform is a function of its compute and memory signature, which in-turn, is governed by a wide range of parameters such as topology size, primitives used, framework used, batching strategy, latency requirements, precision etc. Current benchmarking tools suffer from limitations such as a) being either too granular like DeepBench [1] (or) b) mandate a working implementation that is either framework specific or hardware-architecture specific or  both (or) c) provide only high level benchmark metrics. In this paper, we present NTP (Neural Net Topology Profiler), a sophisticated benchmarking framework, to effectively identify memory and compute signature of an end-to-end topology on multiple hardware architectures, without the need for an actual implementation. NTP is tightly integrated with hardware specific benchmarking tools to enable exhaustive data collection and analysis. Using NTP, a deep learning researcher can quickly establish baselines needed to understand performance of an end-to-end neural network topology and make high level architectural decisions. Further, integration of NTP with frameworks like Tensorflow, Pytorch, Intel OpenVINO etc. allows for performance comparison along several vectors like a) Comparison of different frameworks on a given hardware  b) Comparison of different hardware using a given framework  c) Comparison across different heterogeneous hardware configurations for given framework etc. These capabilities empower a researcher to effortlessly make architectural decisions needed for achieving optimized performance on any hardware platform. The paper documents the architectural approach of NTP and demonstrates the capabilities of the tool by benchmarking Mozilla DeepSpeech, a popular Speech Recognition topology. 
\end{abstract}

\section{Introduction}
Deep Neural Networks are ubiquitous in their deployment to address challenges in Vision and Speech. Neural networks are an area of increased research and development investment with novel end-to-end architectures being developed and deployed across several industry domains. Recently, several organizations are beginning to adapt a 'continuous modeling methodology' where the models are continuously tuned for performance in production environment through an automated-modeling infrastructure. Though there are several frameworks available to build neural net topologies, sophisticated tools to benchmark end-to-end topologies and offer insights for tuning are not available. NTP is an end-to-end benchmarking tool which addresses this gap by enabling detailed benchmarking to understand the compute and memory signature of complete  neural network topology. NTP can be used to  understand compute requirements for a topology as well as to identify compute hotspots, memory bottlenecks etc through run time data flow analysis.

Neural network deployments typically have two phases a) Training and b) Inference. The usual approach to benchmarking is to implement a topology in selected framework and use it for training or inference benchmarking. Inference stage optimization like pruning, quantization etc normally requires retraining. Overall, a time consuming effort. NTP addresses these constraints by enabling a researcher to quickly check the performance impact with different configurations like layer sizing, quantization, pruning etc. NTP is currently targeted to address the benchmarking requirements in inference phase. However, there is no conceptual limitation in the tool preventing its usage in training phase. In addition to compute and memory benchmarking, the tool also allows its users to determine performance metrics like latency, queries per second etc. 

Currently NTP supports Tensorflow (TF), PyTorch, and Intel OpenVINO as underlying framework and allows workloads to run across different hardware platforms like Intel x86 CPU, NVidia GPU, Intel Movidius, Intel GPU and Intel FPGA. NTP does not currently support integration with hardware simulation platforms. NTP allows users to easily construct complex neural-networks as workloads and interface with compatible benchmarking tools for metrics collection. Compute, memory and network bottlenecks can be easily analyzed to enable effective decision making towards optimizing a topology for best performance.

\section{Survey of current profiling tools}
A survey of current profiling tools is presented in this section. Compared to NTP, all these tools lack in more than one area like: a) Lack of ease of model creation b)  Limited support for end-to-end profiling c) High effort pre-requisites like availability of framework/hardware specific implementations d) Lack of support for collecting detailed benchmark metrics e) Lack of support for performance comparison across different target hardware etc.

Certain frameworks like TensorFlow[2] natively supports layer-wise execution-time profiling, but lacks support for extracting detailed benchmark metrics and performance insights. DeepBench[1] is targeted to benchmark neural network libraries (kernels) across different hardware. DeepBench benchmarks common operations for throughput and latency at kernel-level. While kernel-level benchmarks help determine which hardware gives best performance for a chosen kernel, they cannot fully comprehend topology level bottlenecks and hence lack capability to help facilitate topology tuning. Tools like DawnBench[3] and its successor MLPert[4] support end-to-end topology benchmarking for actual implementations of selected ML problems and provides metrics like training and inference and cost. Tools like DLInfBench[5] allows benchmarking of speed and peak memory across frameworks. Again, support is limited to a set of pre-selected topologies as with earlier tools. 

Several of these surveyed tools provide a high-level score for a topology considering training and/or inference. While high level scoring, enables one to compare different topologies and rank them, insights into critical bottlenecks, leading to observed performance, is missing. Also to initiate benchmark for a new topology in these tools, the topology needs to be first created in each framework and supplied to it, which is a resource intensive task. NTP has the advantage of accepting topology definition in a simple, framework and hardware agnostic, format and use it across all frameworks and hardware platforms. The focus of NTP is to provide an exhaustive set of benchmark metrics to help with analysis, identification and resolution of topology and hardware bottlenecks.

\begin{figure}[h!]
\centering
\includegraphics[width=8cm, height=3.5cm]{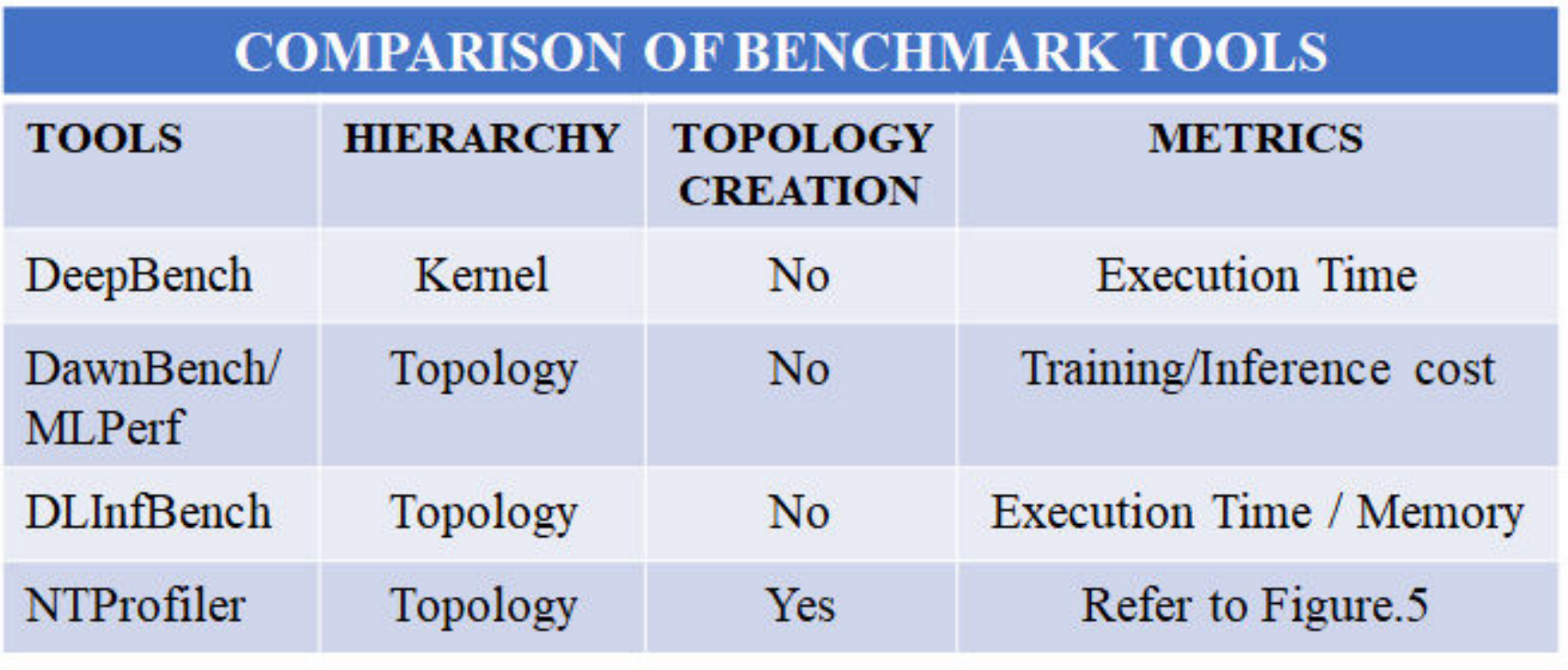}
\caption{Comparison of Benchmark Tools}
\end{figure}

NTP addresses the listed deficiencies by (1) providing a simple markup language based interface for defining neural network (NN) topologies (2) allowing selection of preferred framework (3) simulating the topology on selected hardware platform (including hetero-hardware platforms) and (4) generating detailed benchmark reports for analysis. The implementation details of NTP is presented in the following section.

\section{NTP Overview}
NTP takes the definition of an end-to-end neural network in a simple markup language format, builds the topology corresponding to this definition on the chosen framework and executes the topology by passing data in configured precision through the entire topology. During the execution, NTP collects information on hardware specific performance counters using configured benchmarking tools. NTP leverages the collection capabilities of supported bench tools to observe and summarize performance metrics. Also additional metrics like throughput is also collected to provide detailed insights into parameters that deteriorate performance.

\begin{figure}[h!]
\centering
\includegraphics[width=14cm, height=6cm]{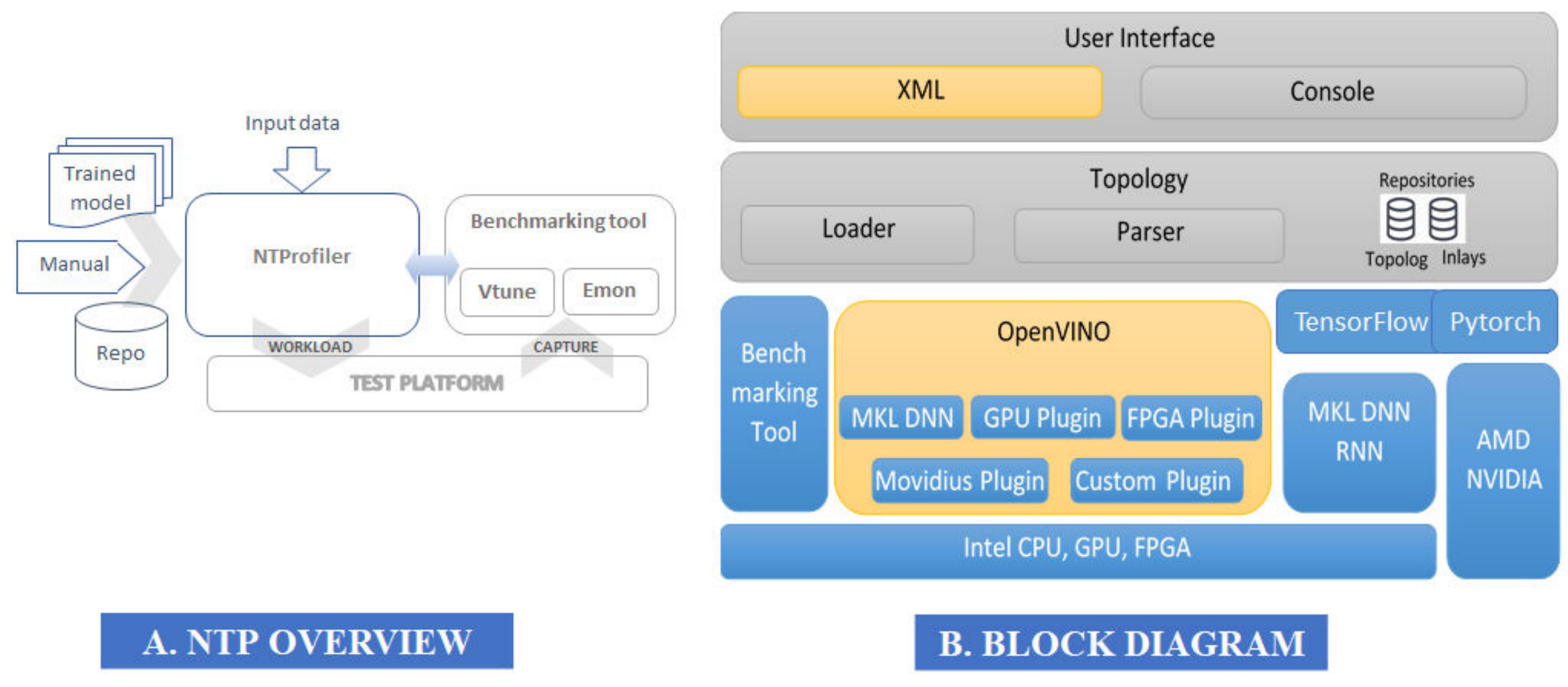}
\caption{NTP Overview}
\end{figure}

A brief description of NTP flow is provided in this section. For processing the input, NTP has a) Loader module and  b) Parser module (Figure 2B). Loader module can directly load a pre-trained model into the configured framework and initiate the benchmark process. In the case that the topology is provided in a XML format, the Parser module will parse the input XML file and generate a neural network graph corresponding to topology description in XML. The format of the generated graph depends on the framework used for profiling. Currently, Tensorflow®, PyTorch® and OpenVino® are supported. It is easy to extend support to additional frameworks. The parser module parses each tag in the input XML file to, obtain attributes of different layers, identify benchmarking markers and builds an  internal graph. The graph is then used to build a model corresponding to the chosen framework. The execution is done in the context of chosen framework and NTP relies on capabilities of the chosen framework for execution.

\subsection{Topology Definition}
To support a wide range of topologies, NTP supports (1) Topologies defined in open formats like ONNX or framework-native formats like Tensorflow pb and (2) Topologies defined using Markup language definition in proprietary XML format. When XML format is used, NTP be automatically builds the model specific to  framework and/or hardware and uses it for benchmarking. Practical neural networks also contain non-neural network functions like MFCC calculation for pre-processing, beam decoders for post-processing, mem copy, format conversions etc. NTP allows these functions also be included as inlays for a realistic end-to-end performance benchmark. Example of a simple topology is shown in Figure 3. Each layer contains one or more primitives like CNN, LSTM etc and inlay functions. NTP provides flexibility to benchmark the entire topology or specify layers-of-interest. For targeted benchmarking, layers-of-interest can be setting using start and end 'markers'. A snippet of the xml corresponding to DeepSpeech is provided in Section 4[Figure 5.B]. 

\begin{figure}[h!]
\centering
\includegraphics[width=10cm, height=4cm]{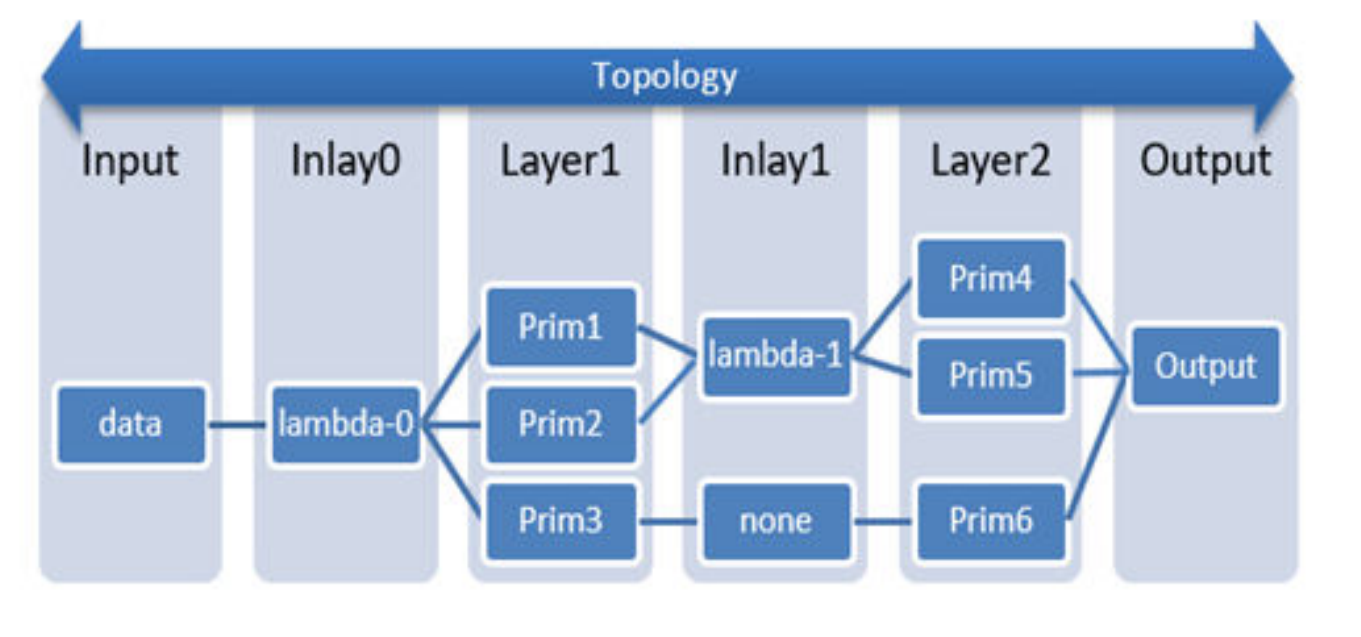}
\caption{NTP Topology Description}
\end{figure}

\subsection{Benchmark Tools and Metrics}
NTP is integrated with a set of benchmarking tools and appropriate tool is chosen based on hardware on which profiling is done. Choice of benchmark tool is also made through marker update in the XML. Currently, tools like VTune, Amplifier etc. offered as part of Intel Parallel Studio, NVidia NSight and command line tools like PCM, EMON etc. are supported. Users can configure and select a specific benchmark tool to be used to determine memory and compute signature of the workload. NTP relies on the ability of selected benchmarking tool to support features like Start, Stop, Pause and Resume etc for targeted metrics collection. The list of metrics supported by Intel Benchmark tool for CPU hardware is listed in Figure 4. NTP automates all the tasks related to benchmarking and generates analysis reports to facilitate quantification of topologies across applicable vectors like layers, topology, frameworks and hardware etc.

\begin{figure}[h!]
\centering
\includegraphics[width=7cm, height=4cm]{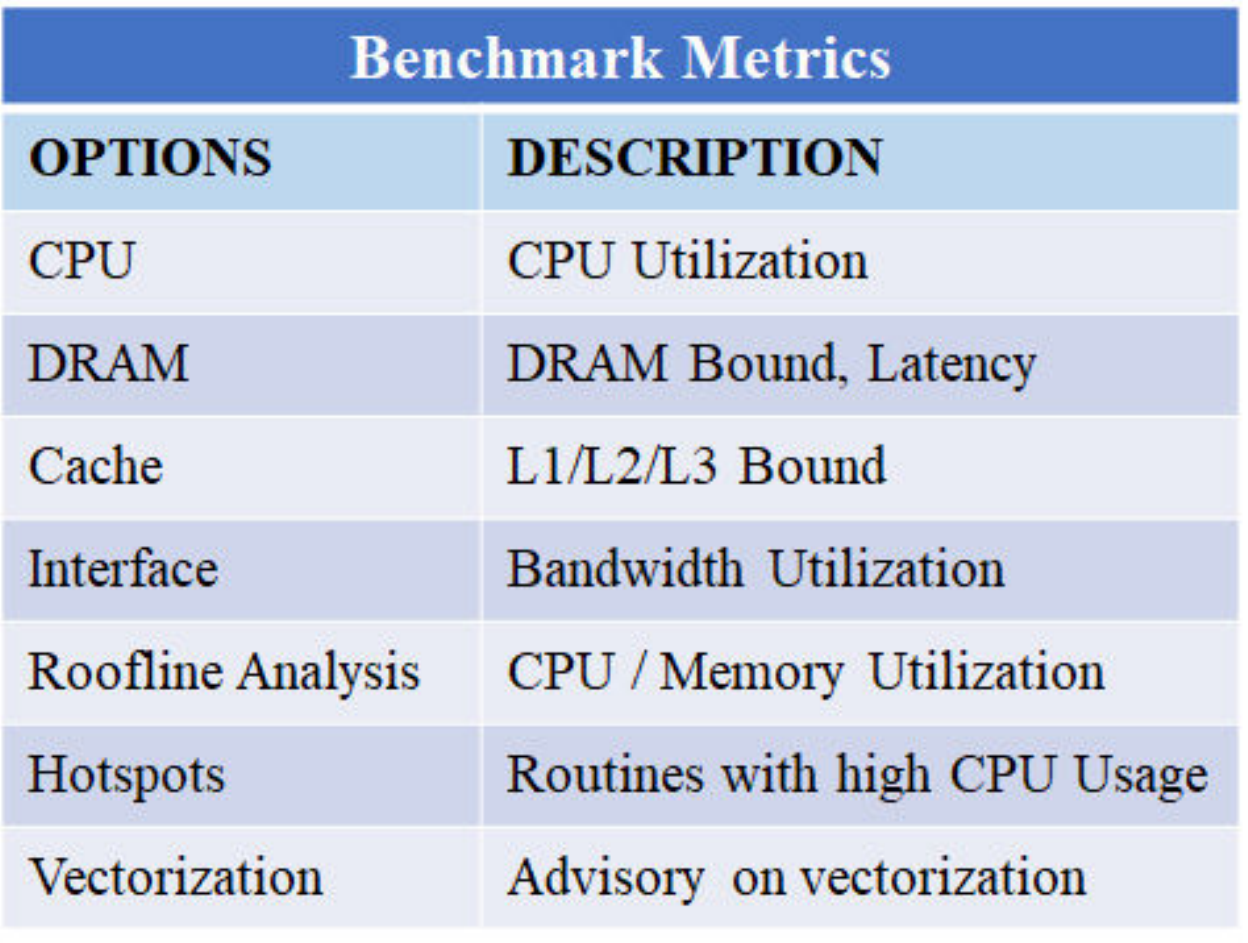}
\caption{Supported Benchmark Metrics}
\end{figure}

\subsection{Framework and Hardware Support}
NTP facilitates topology benchmarking on popular frameworks running on a wide range of hardware platforms without the need for framework/hardware specifc implementation. For instance, when framework chosen is Tensorflow, it leverages TensorFlow’s native hardware support for executing a topology on CPU and NVidia GPU. It is also integrated with Intel OpenVINO framework and can fully leverage heterogeneous compute capability of the framework. OpenVINO currently supports Intel CPU, GPU, GNA, Intel Movidius, FPGA etc. For hardware like Movidius, the support is extended to use a resource-pool of movidius sticks for further acceleration. OpenVINO accepts pre-trained models from popular frameworks like Tensorflow®, caffe® etc. and can perform additional optimization like constants folding, quantization, layer fusion etc to improve performance. In hetero-mode, a workload will allowed to leverage multiple hardware accelerators to meet performance/cost/latency targets. Based on user intent and hardware-support for the constituent kernels, the workload will be automatically partitioned into different subgraphs and each subgraph will be run on its chosen hardware. In addition to eliminating the effort needed to implement a workload for different hardware platforms, OpenVINO also enables NTP to support optimal utilization of available hardware resources. 

Ease of model creation, control over benchmark layers, support for inlays, access to a wealth of benchmark metrics, and support for multiple hardware platforms etc. facilitate users to build, analyze, compare and in-turn optimize neural net topologies in a quick and efficient manner. The capabilites described so far will be demonstrated using a case-study in the following section.

\section{UseCase Mozilla DeepSpeech}
This section demonstrates NTP capabilities as applied to an Automated Speech Recognition workload: Mozilla Deep Speech[6]. DeepSpeech is a character level speech-to-text model that takes Mel Frequency Cepstral Coefficients (MFCC’s) extracted from speech utterance as input and generates textual transcription. The topology has few fully connected layers (FC), bi-directional LSTM (Bi-LSTM) and a final CTC beam search decoder for removing duplicate characters. 

A snippet of input XML for initial few layers of DeepSpeech is presented in Figure 5.B below. As discussed earlier, the choice of XML is primarily to allow users to quickly define a neural network topology and simulate it without the need for framework/hardware specific implementations. The ease of model creation allows for fast iteration over multiple configurations to compare and contrast the hardware implications

\begin{figure}[h!]
\centering
\includegraphics[width=14cm, height=7cm]{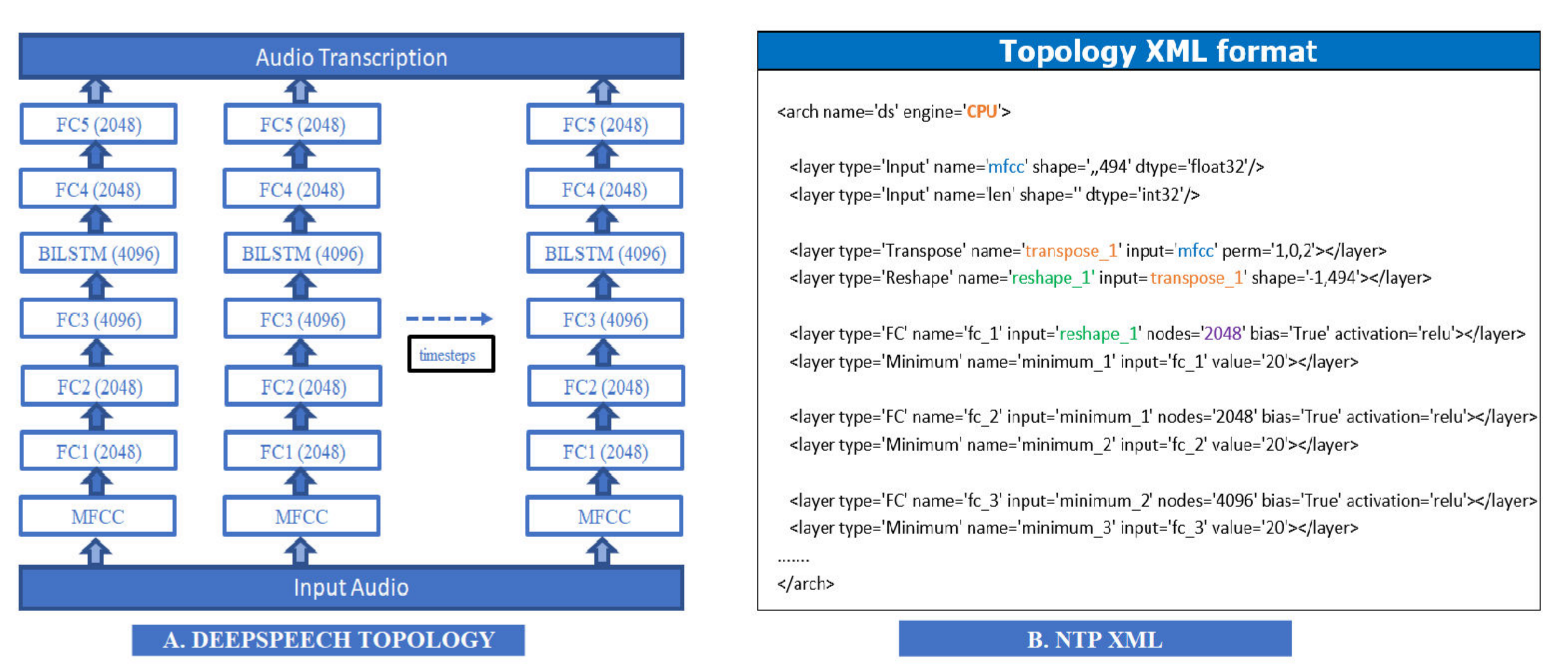}
\caption{Mozilla DeepSpeech}
\end{figure}

From Figure 5.B. above, it can be seen that several topology parameters like layer type, number of nodes for a given layer etc. are all easily specified and updated through the xml. In addition, batching information, data precision, hardware engine, benchmark tool are also accepted as user arguments. Since NTP is a topology exploration and optimization tool, it also supports a topology to executed for performance benchmarking even before actual training. This is done by building a topology model and supplying random weights and biases to the constituent layers. For inputs, dataset in required format (dimension, batch size, precision, range)  is generated with random numbers and fed to the model. Empirical tests have been run to validate that the performance reported by NTP is well aligned with the performance observed from framework-specific implementations of the topology. Figure 6.A shows normalized execution-times measured by running DeepSpeech topology using NTP against execution times seen from framework-specific manual implementation. Minor difference in result is due to extra pre-processing and post processing steps in NTP. For cases where pre-trained model is available, layer weights and biases from the model is directly used.

\begin{figure}[h!]
\centering
\includegraphics[width=12cm, height=5cm]{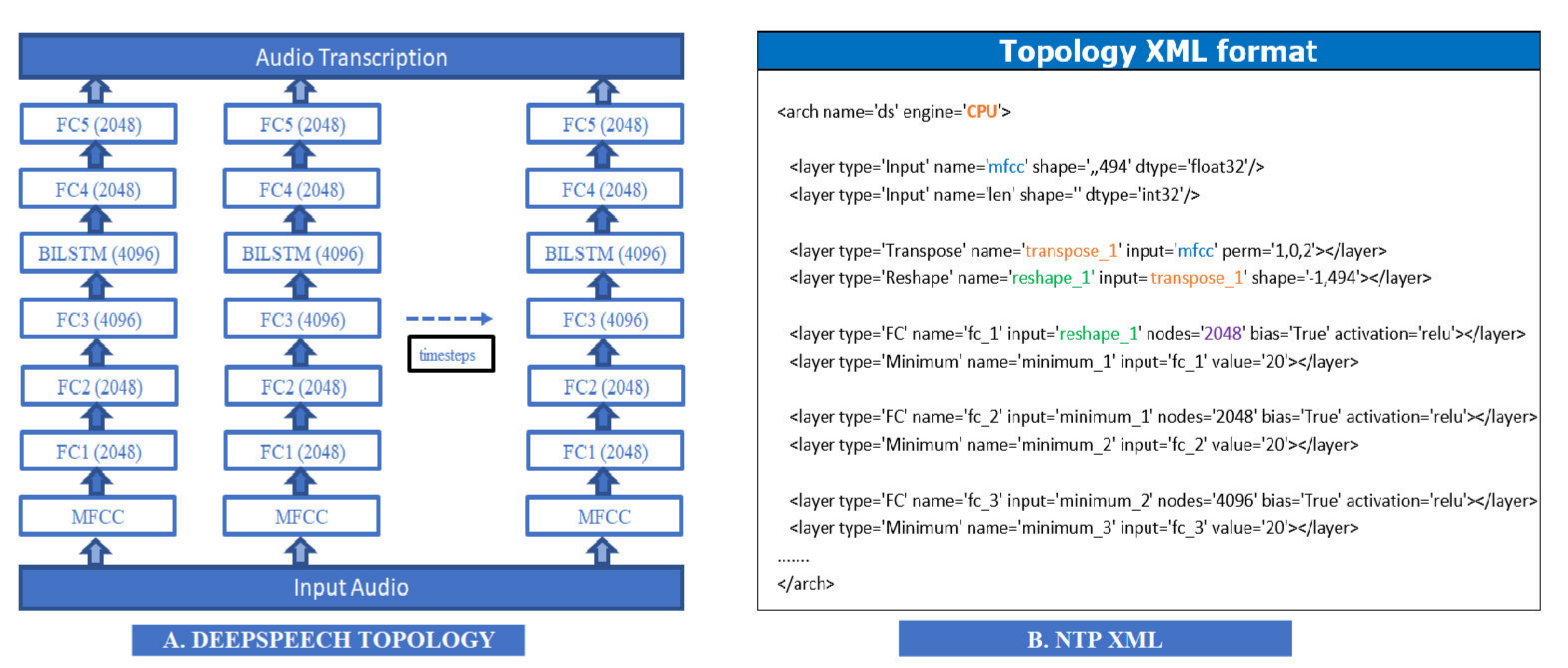}
\caption{A. Comparison of execution times B. DeepSpeech Metrics Summary}
\end{figure}

Due to tight integration with Intel Benchmarking tool-suite for CPU profiling, NTP can provide both high level summary as well as facilitate deep dive into details-of-interest. High level summary of different benchmark metrics for DeepSpeech topology is plotted in Figure.6.B. Depending on kernel-type, kernel-dimensions, input dimensions, memory requirements, cache status etc., each layer runs with a unique execution signature. This is captured and summarized by the tool while data flows through the architecture. Using this information, a tool user can easily a) Optimize the topology to better run the available hardware and/or b) Understand the hardware requirements for a topology and make effective decisions to configure the same. For instance, a kink in CPU time coupled with spike in DRAM (Figure.6.B) is due to memory hungry nature of BI-LSTM layer as will be described in following sections. 

\subsection{Layer Comparison}
Execution times of individual layers of the topology can be summarized and compared using layerwise comparison feature of NTP as shown in  Figure 7.A

\begin{figure}[h!]
\centering
\includegraphics[width=12cm, height=5cm]{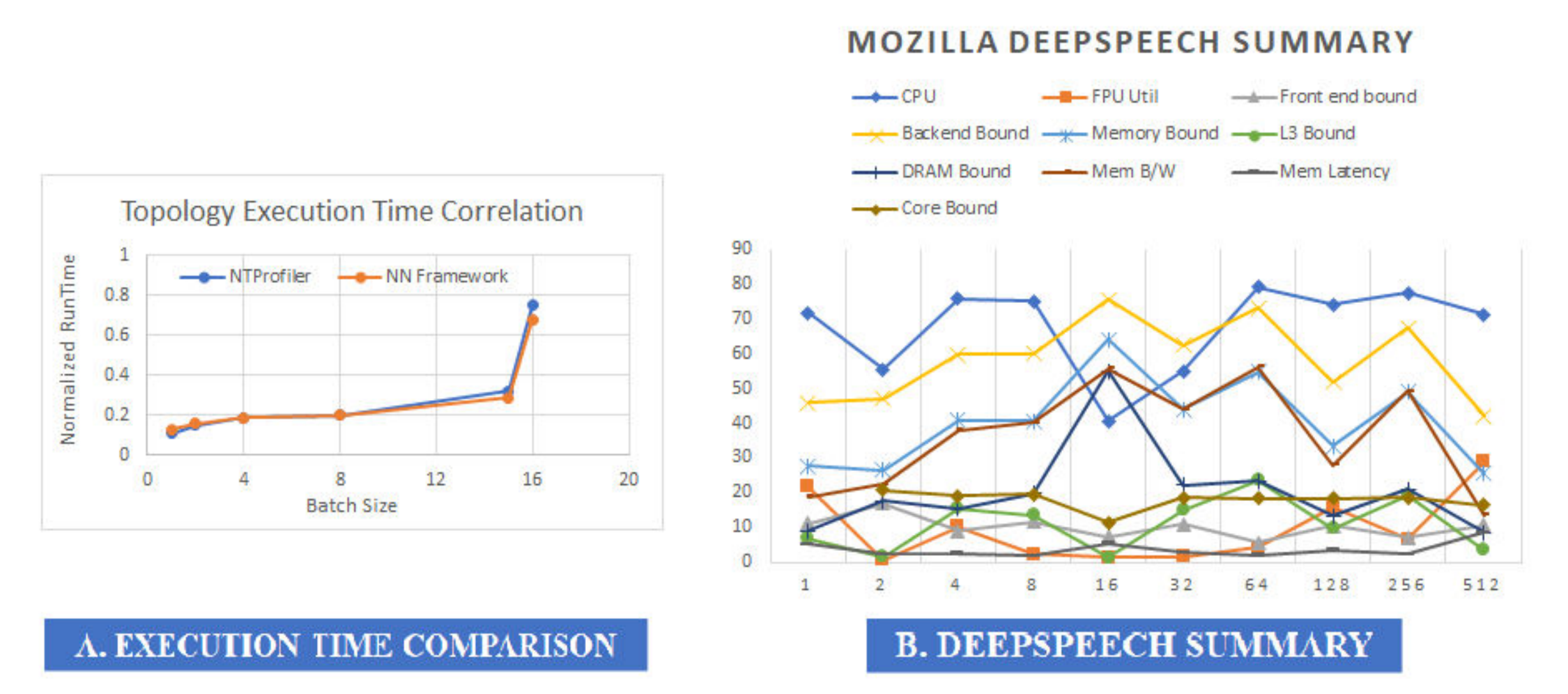}
\caption{Mozilla DeepSpeech Benchmark}
\end{figure}

From above figure, BI-LSTM layer contributes to majority of the runtime. FC layers of this topology do not heavily depend on cache or external memory. However BI-LSTM pulls most of its data/weights from external memory and significantly slows down overall execution. 

In depth analysis of layers can be done using  NTP's targeted benchmarking capabilities. To demonstrate this, the topology was bench-marked as three segments. BI-LSTM was profiled separately and FC layers grouped together  (FC1-3 = FC1+FC2+FC3 : FC4-5 = FC4+FC5). From the benchmark data (Figure 7.B), it can be seen that the FC-Layers are Front-End/Core bound while the BI-LSTM is Backend/Memory bound translating to large runtimes.

\subsection{Topology Comparison}
To quantify the performance impact of parameters like layer sizing, precision etc, relevant parameters can be easily updated in the XML and resultant topologies compared across critical performance metrics. For illustration, DeepSpeech topology[6] was used as reference and number of nodes in different layers were changed to generate two other variants of the topology(Topology1-3 Figure 8.). As described in earlier sections, changes are required only in the input XML, which enables topology modification and hence analysis at an accelerated pace.

As a general disclaimer, the comparison plots presented in this paper are purely to demonstrate NTP tool capabilities. Attention was not paid to benchmarking guidelines or tool configurations for data collection. (Ex: Figure 9.  topology1 is not claimed to be 0.07x faster than topology3 but an observed number for a given run).

\begin{figure}[h!]
\centering
\includegraphics[width=12cm, height=5cm]{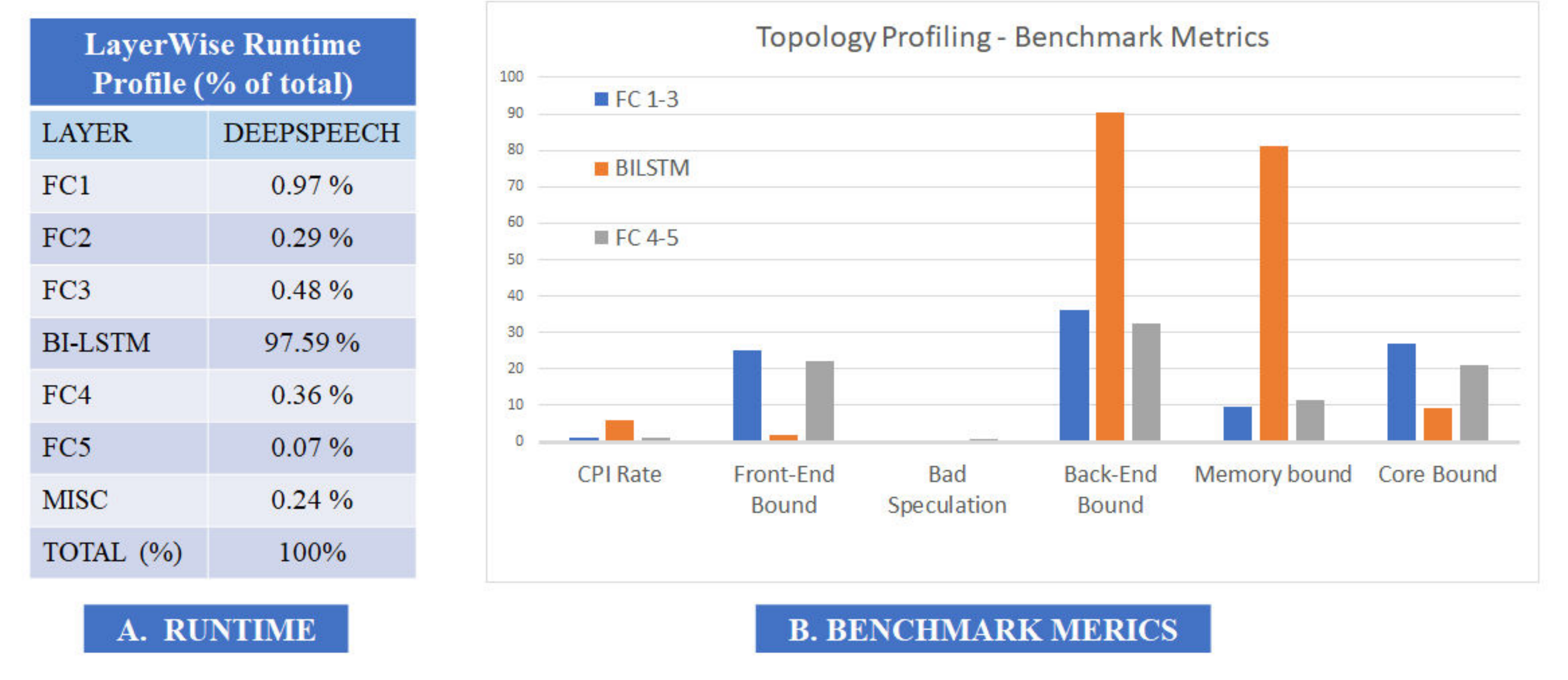}
\caption{NTP Analysis : Topology Comparison - I }
\end{figure}

\begin{figure}[h!]
\centering
\includegraphics[width=12cm, height=6cm]{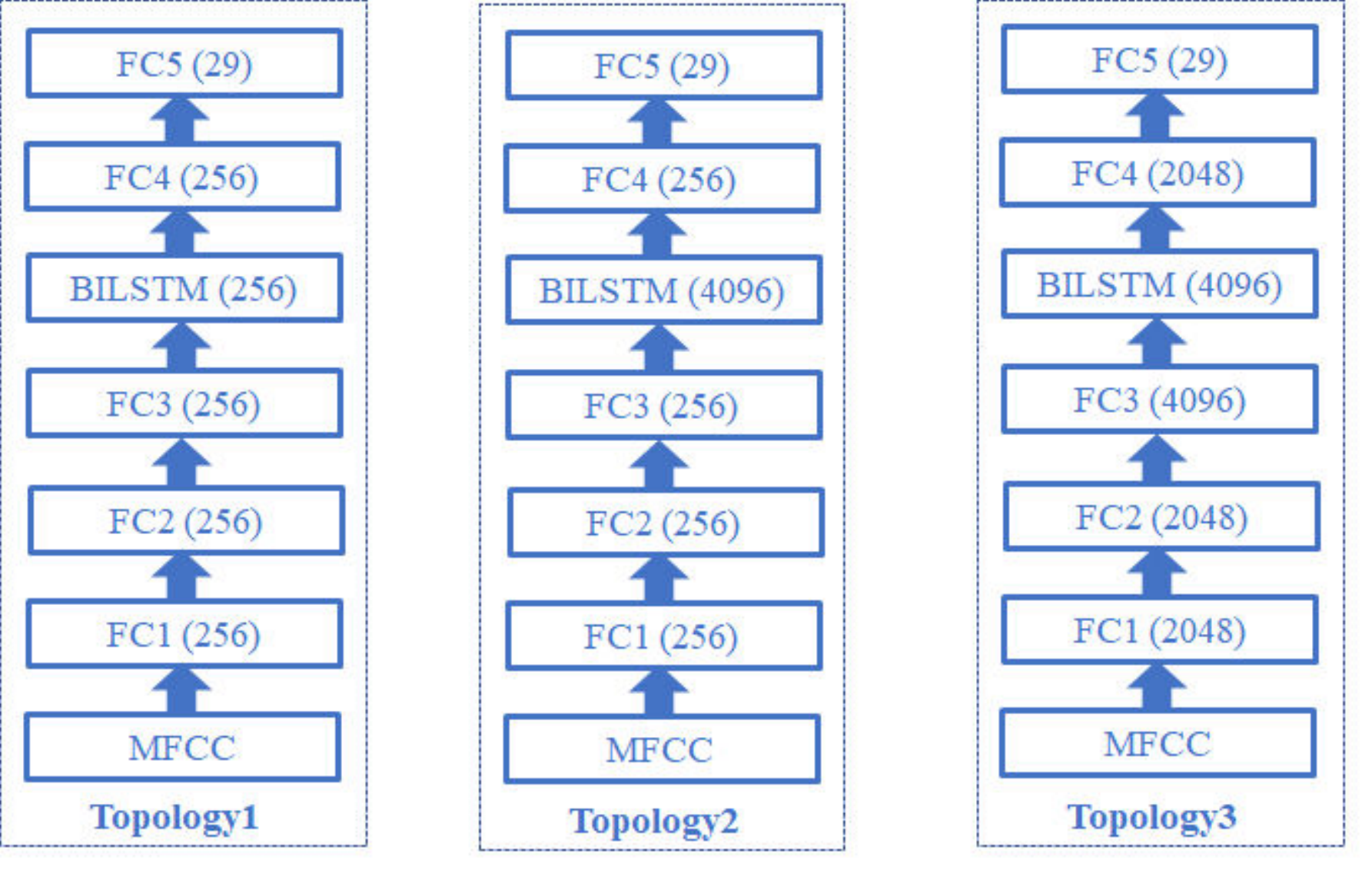}
\caption{NTP Analysis: Topology Comparison - II}
\end{figure}

Percentage run-time contribution from different layers are shown in Figure 9. For topology-1, majority of the time (81.99 \%) is spent in execution of the BI-LSTM layer. The percentage tends to increase further to >97\% if the number of LSTM nodes increases (topology2, topology3). In addition to being memory intensive, BI-LSTM layer cannot be parallelized to the same level as other primitives like CNN. Utilization drops with increase in threads for Bi-LSTM layer due to even higher memory contention leading to further reduction in CPU utilization. As a result, topology would not benefit significantly from multi-threading either. This memory hungry nature of Bi-LSTM seems to be the key performance bottleneck for this topology. And the topology will yield better performance on hardware with better memory capacity and bandwidth.

\subsection{Hardware Comparison}
FPGA's are massively parallel and have lot of on-chip memory and bandwidth when compared across several hardware classes and can be a hardware choice for this topology. To offload execution to a supported accelerator, the choice of hardware alone needs to be updated in NTP run. A comparison of deepspeech runtime signatures on CPU vs FPGA is shown in Figure 10.A below, In addition to significant speedup, LSTM contribution to the total runtime also shows some reduction signifying efficiency.

\begin{figure}[h!]
\centering
\includegraphics[width=12cm, height=6cm]{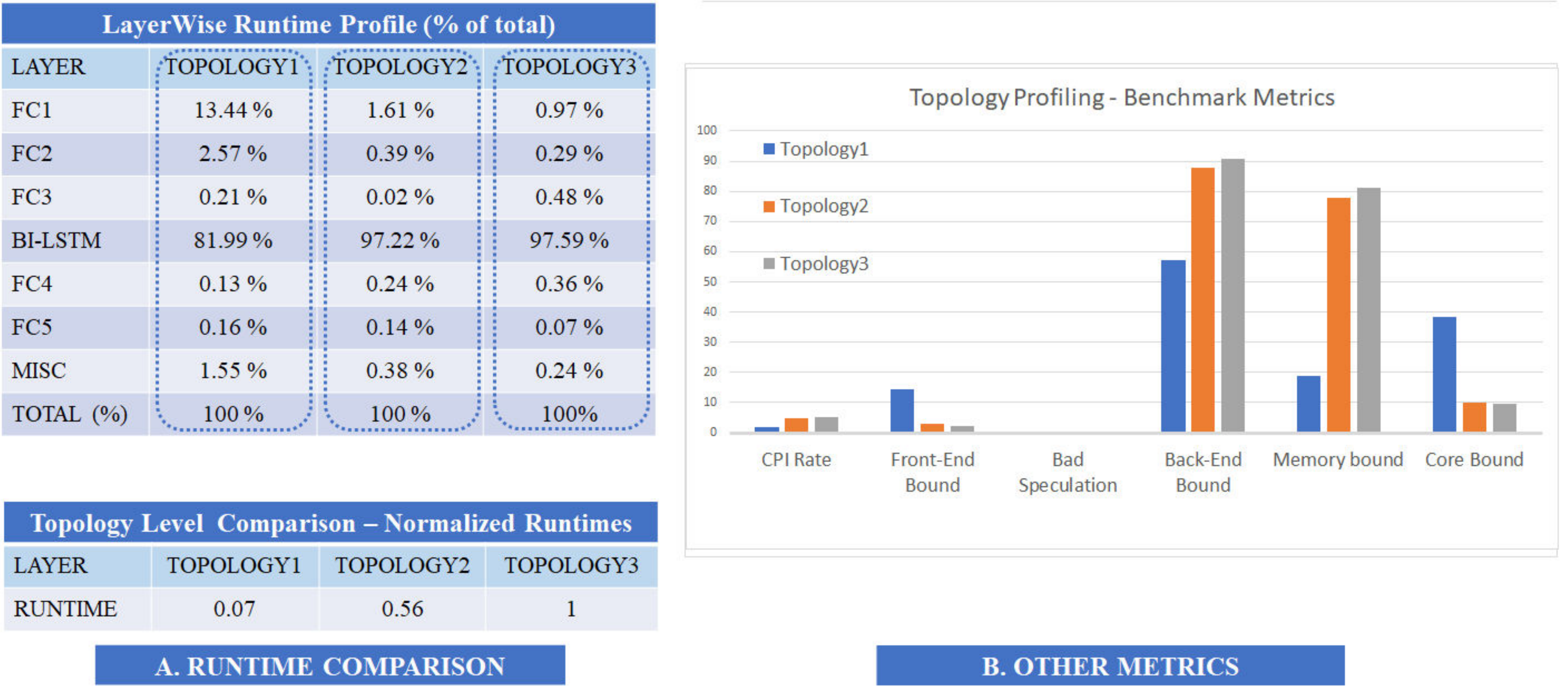}
\caption{NTP Analysis : Hardware and Framework Comparison}
\end{figure}

\subsection{Framework Comparison}
Comparison of performance across frameworks is also a commonly executed benchmark to ensure framework chosen is well optimized for given hardware. NTP allows users to seamlessly switch between supported frameworks. A comparison between execution times for DeepSpeech topology on Tensorflow vs Openvino is shown in Figure 10.B.

Currently, deployments have different training and inference environment due to difference in performance, memory, power, bandwidth, latency requirements of the hardware used for these tasks. Training is typically done in high performance compute farms and inference is mostly done on platforms which can promise real time performance at low cost and power. Topology level analysis helps with identification of  optimal configuration for inference hardware, in-addition to architectural insights for topology optimization. By enabling fast diagnosis of performance/memory bottlenecks with least efforts, NTP aims to help researchers arrive at right set of model parameters and/or hardware configurations at faster pace compared to traditional methods.

\section{Conclusion}
We have presented a topology profiling tool in this paper to help holistically address challenges associated with neural net model development, profiling and tuning. The tool allows for accurate estimation of performance bottlenecks and facilitates quick iterations to optimize the network. We believe this would significantly accelerate the model development, optimization and deployment process for neural network inference.

\section{Future Work}
The current focus of the tool is on inference and the tool can be easily enhanced for training benchmarks as well. NTP is capable of building a topology, perform data collection during execution and reporting. In future, this analysis data can fed directly into an ‘continuous modeling’ environment for targeted tuning. Also the tool can be plugged to a Design of Experiments (DoE) setup to automatically determine best configuration for running heterogenous-compute workloads. 

\section*{References}

\medskip

\small

[1] DeepBench, Baidu. {\it https://github.com/baidu-research/DeepBench}

[2] Martín Abadi, Ashish Agarwal, Paul Barham, Eugene Brevdo,
Zhifeng Chen, Craig Citro, Greg S. Corrado, Andy Davis,
Jeffrey Dean, Matthieu Devin, Sanjay Ghemawat, Ian Goodfellow,
Andrew Harp, Geoffrey Irving, Michael Isard, Rafal Jozefowicz, Yangqing Jia,
Lukasz Kaiser, Manjunath Kudlur, Josh Levenberg, Dan Mané, Mike Schuster,
Rajat Monga, Sherry Moore, Derek Murray, Chris Olah, Jonathon Shlens,
Benoit Steiner, Ilya Sutskever, Kunal Talwar, Paul Tucker,
Vincent Vanhoucke, Vijay Vasudevan, Fernanda Viégas,
Oriol Vinyals, Pete Warden, Martin Wattenberg, Martin Wicke,
Yuan Yu, and Xiaoqiang Zheng.
TensorFlow: Large-scale machine learning on heterogeneous systems,
2015. Software available from tensorflow.org. {\it https://www.tensorflow.org}

[3] DAWNBench, Stanford University {\it https://github.com/stanford-futuredata/dawn-bench-entries}

[4] MLPerf, Baidu. {\it https://mlperf.org/}

[5] DLInfBench. {\it https://github.com/nicklhy/DLInfBench}

[6] Awni Hannun, Carl Case, Jared Casper, Bryan Catanzaro, Greg Diamos, Erich Elsen, Ryan Prenger, Sanjeev Satheesh, Shubho Sengupta, Adam Coates, Andrew Y. Ng. {\it arXiv preprint arXiv:1412.5567}, 2014.

\end{document}